\documentclass[11pt]{article}
\usepackage{nodalida2025}
\usepackage{times}
\usepackage{url}
\usepackage{latexsym}
\usepackage{graphicx}
\usepackage{todonotes}
\usepackage{booktabs}
\usepackage{longtable}
\usepackage{verbatim}
\usepackage{enumitem}
\usepackage{multirow}
\usepackage[hidelinks]{hyperref}
\setlist[enumerate]{itemsep=-1.5mm}

\aclfinalcopy % Uncomment this line for the final submission

\title{Quantification of Biodiversity from Historical Survey Text\\ with LLM-based Best-Worst Scaling}

\author{Thomas Haider, Tobias Perschl, Malte Rehbein\\
  Chair of Computational Humanities \\
  University of Passau \\
  {\tt firstname.lastname@uni-passau.de}} 

\date{}

\begin{document}
\maketitle
\begin{abstract}
  %In this study, we evaluate methods for estimating 
  %species' occurrence frequencies 
  %from historical survey texts through quantity estimation. 
  In this study, we evaluate methods to determine 
  %biodiversity 
  %relative 
  the frequency of species via quantity estimation from historical survey text.
  To that end, we formulate classification tasks and finally show that this problem can be adequately framed as a regression task using Best-Worst Scaling (BWS) with Large Language Models (LLMs). 
  We test Ministral-8B, DeepSeek-V3, and GPT-4, finding that the latter two have reasonable agreement with humans and each other.
  We conclude that this approach is more cost-effective and similarly robust compared to a fine-grained multi-class approach, allowing automated quantity estimation across species. 
  %We find that ... and our case study shows that ... concluding that ...
\end{abstract}

%\section{Calls}

%Main Conference: \\ \url{https://www.nodalida-bhlt2025.eu/conference} \\
%EcoNLP Workshop: \\ \url{https://econlpws2025.di.unito.it/} \\

%\section{Overview}

\iffalse
\begin{enumerate}
    \item Intro: Problem Description
    \begin{itemize}
        \item Heute: Fragebögen, Citizen Science, z.B. Rote Liste (Datenaggregation)
        \item Historical Data: First (systematic) attempts at cataloging biodiversity through surveys
        \item Classification of Biodiversity from Text
        \item Automatically assess/extract the data 
        \item Heterogeneous text data, non-standardized expressions of frequency
        \item Quantify actual biodiversity from textual markers
        \item Identify robust features/classifiers for classification/regression to estimate quantity
        \item 
    \end{itemize}
\end{enumerate}
\fi

%***************************
%       INTRODUCTION
%***************************
\section{Introduction}
%\cite{RS:13} \\
%\citet{RS:13} showed that x.\\
%It was shown that x \citep{RS:13}. \\

Long-term observation data plays a vital role in shaping policies for preventing biodiversity loss caused by habitat destruction, climate change, pollution, or resource overexploitation \citep{dornelas_biodiv, hoque2024addressing}.
%\textcolor{purple}{Monitoring and accurate measurement of biodiversity over extended periods} provide invaluable insight into ecological trends.
However, these efforts depend on the availability of reliable and relevant historical data and robust analytical methods, a significant challenge due to the heterogeneity of records representing such data.

The available biodiversity data varies widely 
in resolution, ranging from detailed records (e.g., point occurrences, trait measurements) to aggregated compilations (e.g., Floras, taxonomic monographs) \cite{konig2019biodiversity}. Many projects, such as the \textit{Global Biodiversity Information Facility} (GBIF), focus largely on the disaggregated end of the spectrum, particularly with presence/absence data \cite{dorazio2011modern,iknayan2014detecting}. 
Furthermore, despite their utility, longitudinal 
data is largely confined to records from after 1970 \citep{Van_Goethem2021-z}, leaving significant historical gaps. 

Natural history collections and records from the archives of societies present valuable opportunities to extend data further back in time \citep{Johnson2011-ah, Bronnimann2018-jg}. 
Such sources are rich, but typically unstructured and require sophisticated extraction tools to produce meaningful quantitative information.
%These sources, often rich in unstructured data like text or images, require sophisticated extraction tools to produce meaningful insights. 
Recent advances in NLP have shown promising potential for retrieval-based biodiversity detection from 
(mostly scientific) literature \citep{Kommineni2024-ne,langer2024relation,lucking2022multiple}.

%Jede Zeile extra hier zerschießt das ganze Paper.
% https://arxiv.org/abs/2311.04929

%Despite its significance, longitudinal biodiversity data is typically confined to post-1970s sources \citep{Van_Goethem2021-zl}, leaving significant historical gaps. Historical sources such as natural history collections and records from the archives of societies offer valuable opportunities for extending datasets further back in time \citep{Johnson2011-ah, Bronnimann2018-jg}. 
%Such (text or image-based) sources 
%are rich in data but typically unstructured. This requires sophisticated extraction tools to produce meaningful insights from quantification. Initial research in this area leverages recent advances in NLP methods, enabling information retrieval based biodiversity detection in (scientific) literature \citep{Kommineni2024-ne,langer2024relation,lucking2022multiple}.

This paper focuses on evaluating methods for biodiversity quantification from 
semi-structured historical survey texts.
To achieve this, we 
test tasks to distill meaningful metrics from  textual information found in survey records. 
A particular focus lies on the feasibility of Best-Worst Scaling (BWS) with a Large Language Model (LLM) as an annotator, which promises
greater efficiency and cost-effectiveness compared to manual annotation
\cite{bagdon2024you}.
%\textcolor{red}{Diversity vs. Endangerment Status}
%Research Questions:}
In the following, we describe the data,
%and datafication process 
%(section~\ref{sec:data}),
outline the 
%operationalization of the 
tasks and machine learning methods, %(section~\ref{sec:tasks}),
and finally present a case study. 
\section{Data}\label{sec:data}

\begin{table*} 
\setlength{\aboverulesep}{1pt} % Reduce space above \midrule
\setlength{\belowrulesep}{2pt} % Reduce space below \midrule
    \centering\scriptsize
    \begin{tabular}{llccll}%{l|l||l|l||l|l}
    \toprule
    \textbf{Animal} & \textbf{Text} & \textbf{Binary} & \textbf{BWS} & \multicolumn{2}{l}{\textbf{Multi-Classification}}\\
    \midrule
        Ducks & Bedecken Isar-Strom, wie Amper und Moosach in ganzen Schwärmen. & 1 & 1.00 & 5 & \textsc{Abundant}\\ & \textit{Cover  Isar-stream, likewise Amper and Moosach in whole swarms.} &&& \\ \midrule
        %Eurasian Otter & Ziemlich häufig an der Altmühl und Laber & 1 & 0.69 & 5 & Abundant\\ & Quite common at the Altmühl and Laber (rivers). &&& \\ \hline
        Roe Deer & Ist hier zu Hause, und beinahe in allen Waldtheilen zu finden. &1 & 0.88 & 4 & \textsc{Common} \\ & \textit{Is at home here and can be found in almost all parts of the forest.} &&& \\ \midrule
        European Adder & Kommt wohl aber eben nicht häufig vor. & 1 & 0.44 & 3 & \textsc{Common to Rare}\\ & \textit{Does indeed appear but just not that often.} &&& \\ \midrule
        Lynx & Höchst selten wechseln derlei Thiere von Tyrol herüber.& 1 & 0.12 & 2 & \textsc{Rare} \\& \textit{Very rarely do such animals cross over from Tyrol.} &&& \\ \midrule
         Wild Goose & Kommt nur äußerst selten zur Winterszeit vor.& 1 & 0.06 & 1 & \textsc{Very Rare} \\ & \textit{Occurs only very rarely at winter time.} &&& \\ \midrule
        Owl & Horstet dahier nicht und verstreicht sich auch nicht in diese Gegend. & 0 & 0.00 & 0 & \textsc{Absent} \\ & \textit{Does not nest here and does not stray into this area.} &&& \\ \midrule
        %Beaver & Um Passau nicht, bey Landshut sah ich 1826 im Freyen noch wilde Biber. & 0 & 0 & -1 & Extinct \\ & Not around Passau, near Landshut I did see wild beavers out in the open in 1826. &&& \\ \hline
        Wolf & Kommt nicht mehr vor. & 0 & 0.00 & -1 & \textsc{Extinct} \\
        & \textit{No longer occurs.} &&& \\ 
        \bottomrule
    \end{tabular}
    \caption{Data Examples with Annotation (our own translations)}
    \label{tab:data_example}
\end{table*}

In 1845, the Bavarian Ministry of Finance issued a survey to evaluate biodiversity in the Bavarian Kingdom, a region that encompasses a variety of different ecosystems and landscapes. To that end, 119 forestry offices were contacted to complete
%and return 
a standardized questionnaire. Namely, trained local foresters recorded in free text 
%whether and 
how frequently 44 selected vertebrate species occurred in the respective administrative territory, and in which habitats and locations 
%of the respective administrative territory 
they could be found. 

\begin{figure}[htbp!]
    \centering
    \includegraphics[width=.35\linewidth]{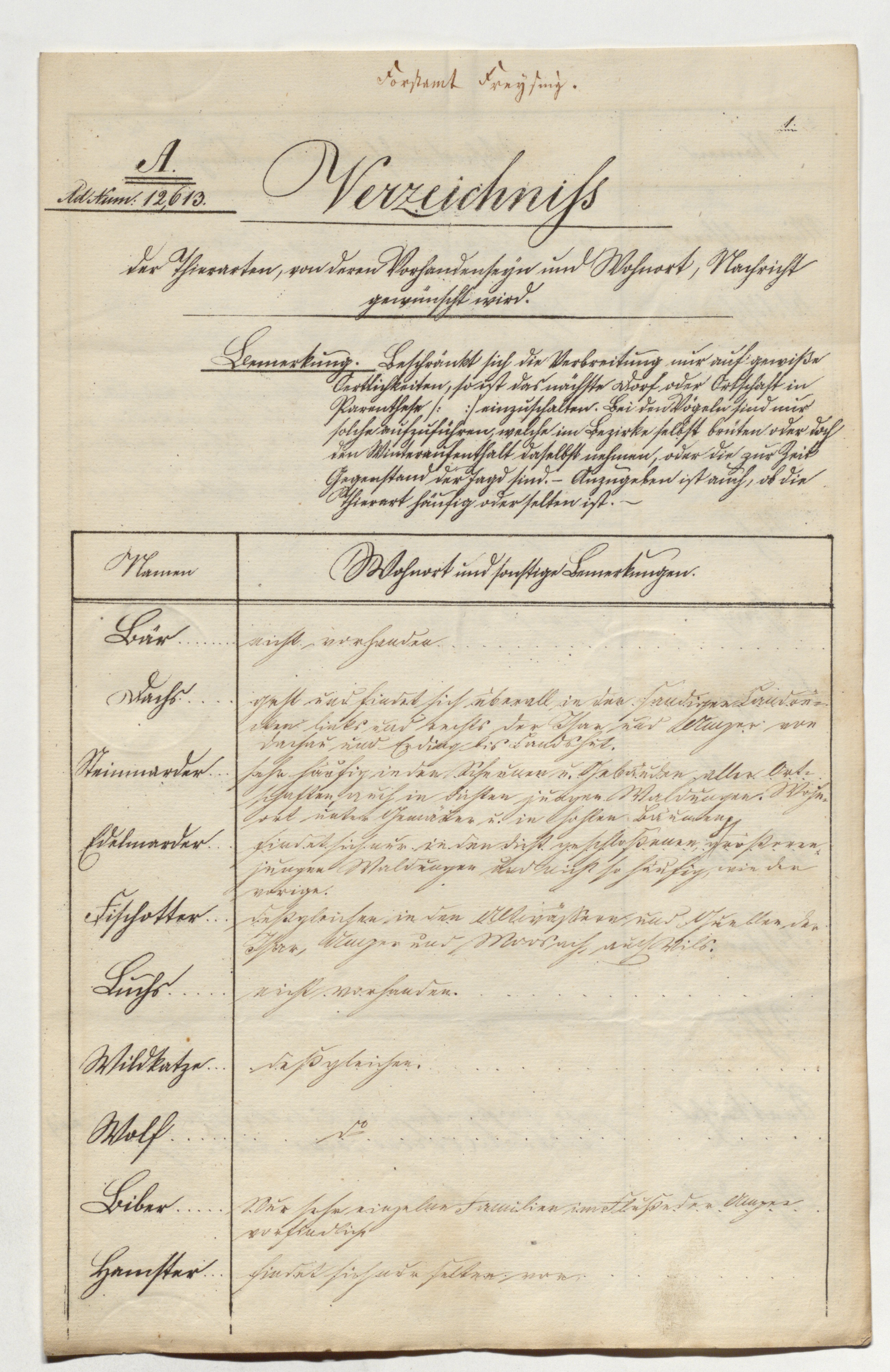}
    \caption{Facsimile of a survey page, Freysing forestry office in the Upper Bavaria district.}% from the Freysing forestry office in the Upper Bavaria district, with a header containing instructions and a number of records describing animal species with their respective responses. These historical documents are preserved by the Bavarian State Archives \citep{rehbein2024historical}.}\todo{Bitte die Captions überschaubar halten. Text kommt in den Artikel, nicht in die Caption. Bayr. Staatsarchiv muss trotzdem zitiert werden!}
    \label{fig:questionnaire}
\end{figure}

Figure~\ref{fig:questionnaire} shows the facsimile of a digitized survey page. It features a header containing instructions and a number of records describing animal species with their respective responses. These historical survey documents are preserved by the Bavarian State Archives \citep[cf.][]{rehbein2024historical}.

The archival sources were digitized, transcribed
%\footnote{With the aid of Transkribus (\url{transkribus.org})} 
from the handwritten original and enriched with metadata, including, among others, taxonomic norm data according to the GBIF-database\footnote{\url{gbif.org}}  \citep{telenius2011biodiversity}
%,ivanova2021possibilities}
%\footnote{For the purposes of this study, we consider 44 selected vertebrate species.}  
and geographical references to forestry offices. 
%The transcription of the documents was carried with the aid of Transkribus\footnote{\url{transkribus.org}} handwritten text recognition. 
This data set is freely available on Zenodo \citep{rehbein2024historical}.
%: \url{https://doi.org/10.5281/zenodo.14008158}

In total, the data set contains
%\todo{laut meiner rechnung is 119x44=5236}
%over 
5,467 entries\footnote{
%One text entry per species and office, but also 
Including species that were not explicity prompted.}
among which are also a number of empty (striked out) or `see above'-type responses. The unique set we used for our experiments contains 2,555 texts.
%As can be seen in Figure \ref{fig:Tokencount}, 
We find that the foresters' replies vary considerably in length where most texts contain 3 to 10 tokens and only a few texts more than 20 tokens. Table~\ref{tab:data_example} provides examples with annotation according to the tasks detailed in the next section.

%\begin{figure}[htbp!]
%    \centering
%    \includegraphics[width=.5\linewidth]{figures/Token_counts_per_text.png}
%    \caption{Histogram of Token Counts in the Texts}
%    \label{fig:Tokencount}
%\end{figure}

%***************************
%       TASKS
%***************************
\section{Tasks \& Experiments}\label{sec:tasks}

The main task in this paper is to assign a quantity label to a text, indicating the 
frequency 
%abundance
with which an animal species occurs in a specific area.
This can be operationalized in various ways, either through a classification task or through regression. In both, it can be as difficult to obtain consistent labels by asking humans to assign a value from a rating scale \citep{schuman1996questions, likert1932technique}.
Likewise, it is also difficult for researchers to design rating scales, 
%as there are many 
considering design decisions such as scale point descriptions or granularity may bias the annotators.
%as it is for researchers to develop such a scale, 
%as there are many 
%considering that design decisions such as scale-point descriptions or granularity may bias the annotator.
%, such as scale point descriptions and scale granularity.
%However, classification systems often introduce a degree of arbitrariness, and as the complexity of the task increases, so does the task of annotation. 
%Regression addresses the issue of arbitrary classes, but annotating for continuous value labels comes with its own set of challenges.
%To operationalize quantity estimation in our data, w

We evaluate three different task setups,\footnote{Code: \url{github.org/maelkolb/biodivquant}}
%\footnote{\url{github/Maelkolb/EcoNLP_Quantification}} 
as detailed in Table~\ref{tab:data_example}: Binary 'Presence vs. Absence' Classification, a 7-ary Multi-Class setup (Abundant to Extinct), and continuous values scaled to $[0,1]$. For the first two tasks, we use manual annotation, while continuous values are derived through BWS with LLMs \citep{bagdon2024you}.
%, a method that is cognitively and computationally more robust than naive annotation \citep{bagdon2024you,kiritchenko-mohammad-2017-best}. 

%the continuous values are derived through manual annotation of quantifiers, and best-worst-scaling with GPT-4, 

%\subsection{Binary `Presence vs. Absence'}
\subsection{Binary Classification}
%We consider t
The simplest form of animal occurrence quantification is a binary distinction between the absence (0) or presence (1) of a given species, an annotation scheme as popular as it is problematic in biodiversity estimation.\footnote{Since \textsc{Absence} may just stem from non-detection, rather than real absence \cite{dorazio2011modern,iknayan2014detecting,kestemont2022forgotten}.}
In our annotation, the label \textsc{Present} is given when a species is described in the historical dataset as having been observed in that particular locality at the time of the survey (thus excluding mentions of past occurrences, i.e., extinctions).
%In our annotation, the label \textsc{Present} is given when a species was described in the historical data set as been observed in that particular locality at the time of the survey (thus excluding 
%explicit 
%mentions of past occurrence, i.e., extinction). 
%Specific types of appearance (e.g., migration patterns) were only considered as appearance. 
The annotation workflow consists of
%multiple
iterative steps with discussions. Agreement is nearly perfect. Overall, from the set of 2,555 unique texts, 1,992 (78\%) fall into class \textsc{Present}, 563 (22\%) into \textsc{Absent}.\footnote{In the complete dataset, absence texts make up more than half of all text descriptions, but often amount to empty or `strike-out' responses. Thus, the task would be easier on the full dataset, because many instances are trivial to predict.}
%, and the majority baseline is lower.}
%In comparison, i

%\subsubsection{Presence/Absence Models}

To test the feasibility of the binary task, we create training curves with different models, namely BERT against 
%baselines with 
Logistic Regression, SVM, and Random Forest on Unigrams. We use 20\% of the data for testing, and take another 20\% from the training set for 
%validation and 
hyperparameter search at each cumulative 100 text increment. Despite the 78\% majority baseline, we find that the models perform well and training requires only a few hundred texts to reach an F1-macro score in the high 90s. 

\begin{figure}[htbp!]
    \centering
    \includegraphics[width=.65\linewidth]{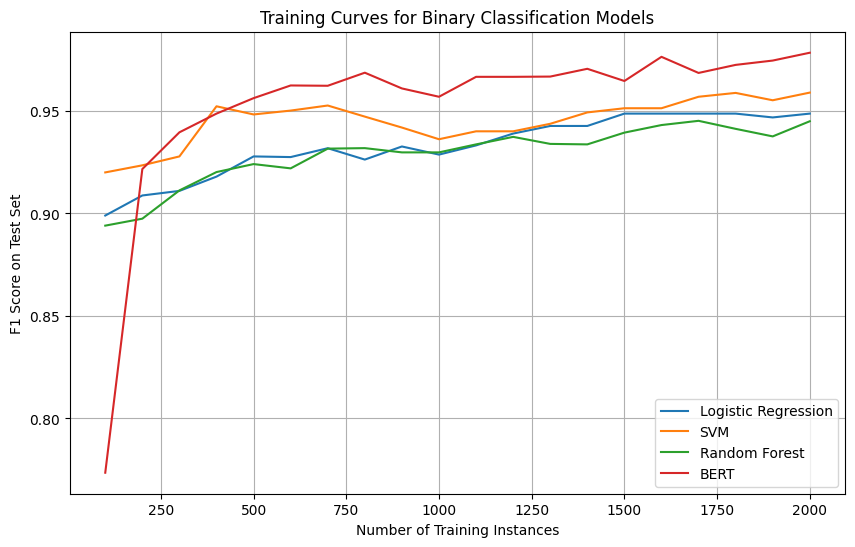}
    \caption{Training Curves of different models on incremental training data (binary classification)}
    \label{fig:trainingcurves}
\end{figure}

Upon feature weight interpretation of the Logistic Regression and LIME on BERT \citep{ribeiro2016should}, we find that there is some bias in the data: Classification decisions occur on tokens that are not explicit quantifiers and easily substitutable without changing the classification result (e.g., common toponyms such as `Danube'). This presents a case of spurious correlations---an interesting future research direction, but a matching \cite{wang-culotta-2020-identifying} or counterfactual approach \cite{qian-etal-2021-counterfactual} appears 
%to be 
challenging for this heterogeneous data. Yet, we annotate the best features 
%weights 
with regard to their `spuriousness' and find that classifiers are still robust without spurious features. This annotation also gives us a list of quantifiers which we utilize for transfer learning of a regression model (section~\ref{sec:regression}).

\subsection{Multi-Classification}

Since the quantification 
of species 
%abundance
frequency 
in practice exceeds the binary differentiation between presence and absence of animals, a multi-class approach provides more details.
%on the actual amount of animal occurrence. 
We use a 7-class system, categorizing texts based on the schema as shown by the descriptors in Table~\ref{tab:data_example}, ranging from \textsc{Abundant} (5) to \textsc{Extinct} (-1).
%, Common (4), Common to Rare (3), Rare (2), Very Rare (1), Absent (0), and Extinct (-1). 
We decided to annotate data of four species for our case study (section~\ref{sec:discussion}): Roe deer, Eurasian Otter, Eurasian Beaver, Western Capercaille, each within the 119 forestry offices (with one annotator).
%(thus, 119 texts per species)
%Roe deer (Capreolus capreolus), Eurasia Otter (Lutra lutra), Eurasia Beaver (Castor fiber) and Western Capercaille (Tetrao urogallus).\footnote{All Latin classifications according to \citet{linnaeus1758systema}}}

A second person annotates a random sample of 100 texts, resulting in a Cohen’s $\kappa$ of 0.78, indicating high agreement.

%We had a second person annotate a random sample of 100 texts which results in a Cohen's $\kappa$ of 0.78, indicating high agreement. 
%on this relatively small sample. 
%\todo{hier hab ich agreement hinzugefügt}

We then train a few models with a 5-fold cross validation, and find that the language agnostic sentence encoder model LaBSE  \cite{feng2022languageagnosticbertsentenceembedding} performs better than monolingual BERT-models and a Logistic Regression. We also test a zero shot classification with GPT-4 and Deepseek-V3. See appendix for the prompt.
%For the logistic regression baseline, we utilized grid search for parameter tuning. 

\begin{table}[htbp!]
\centering\scriptsize
\begin{tabular}{lcc}
\toprule
\textbf{Model} &  \textbf{F1 Micro} & \textbf{F1 Macro}  \\ \midrule
Logistic Regression     & 0.69 &0.61  \\ 
gbert-base   &0.63 &0.51 \\ 
bert-base-german  &0.73& 0.63  \\ 
LaBSE    & \textbf{0.77} & \textbf{0.68} \\ 
GPT4 Zero Shot & 0.70 & 0.56 \\ 
DSV3 Zero Shot & 0.66 & 0.66 \\ \bottomrule
\end{tabular}%
\caption{Multi-class model performance.}% evaluation, 5-fold cross validation}
\label{tab:multi_models}
\end{table}

As seen in Table~\ref{tab:multi_models}, this task is generally quite challenging. 
We find that the main problem is posed by the underrepresented classes, as shown by the discrepancy between micro and macro scores,
%as seen in Figure \ref{fig:multi-class_perclass} 
indicating that more data would help, which is, however, expensive to obtain. Zero shot classification with GPT-4 in turn is biased towards the \textsc{Rare} 
%frequencies, 
classes,
such that \textsc{Common} categories are harder to predict, while DeepSeek-V3 (DSV3) shows a more balanced response.
%We explore the use of large language models for this 7-class annotation. The prompt contains instructions, such as detailing the annotation scheme, or specifying the language of the texts as german.

\iffalse
\begin{figure}
    \centering
    \includegraphics[width=0.5\linewidth]{figures/multi_class_perclass.png}
    \caption{Caption}
    \label{fig:multi-class_perclass}
\end{figure}
\fi 

%, as seen in \textcolor{red}{Table X}.

%\textcolor{red}{More explanation/interpretation.}

\subsection{Continuous Quantification}\label{sec:regression}

Finally, we experiment with operationalizing our task as a regression problem with the aim of generalizing the quantification problem to less arbitrary categories
%with superior adaptability to fine-grained labels 
and a possibly imbalanced data set \citep{johan-berggren-etal-2019-regression}. While a na\"ive labeling of quantifiers showed promising results, it is a challenge 
%(and subjective) 
to create a comprehensive test set based on heuristic annotation. Thus, we experiment with Best-Worst Scaling, aided by LLMs.

%, which is desirable for a number of reasons: 1) Regression allows us to model continuous trends in animal occurrence, capturing granular variations that classification could miss. 2) Regression models demonstrate superior adaptability to fine-grained labels and imbalanced datasets, as supported by evidence from automated essay scoring research \citep{johan-berggren-etal-2019-regression}. 3) Continuous labels generated through a text ranking approach, as detailed in the following section, can be considered less arbitrary compared to the categorization of texts into predetermined classes. 
%We utilized two approaches of training data creation for continuous labels; firstly, we manually extracted token and terms that indicate some form of quantification, scaling them from 0 to 1 manually. By calculating the mean value of all quantifier present within a text, we generate training data for the entire corpus. However, since these quantifier scores are clouded by subjectivity, the language processing capabilities of LLM's can provide an unbiased solution to the ranking of texts in order to calculate objective continuous target labels. \todo{hier habe ich was eingefügt, passt das so?}

\subsubsection{Best-Worst Scaling with LLMs}

%Humans are fairly bad at assigning numerical values to text. However, comparison tasks are cognitively fairly easy.  
%The decision to utilize a best-worst-scaling approach of texts instead of a Likert scale \citep{likert1932technique} or a pairwise comparison setting lies in several factors. Essentially, Likert or other kinds of rating scales "do not force respondents to discriminate between items"\citep{BWS_Theory} possibly causing the scale to contain clusters of items with "similarly high importance"(ibid.) instead of a similar distribution of values across the entire scale. 

%Yet, exhaustive pairwise comparisons are expensive. 
%For instance, comparing 1000 texts pairwise to one-another would take half a million iterations. 
Best-Worst Scaling (BWS) is a comparative judgment technique that helps in ranking items by identifying the best and worst elements within a set. This approach is easier to accomplish than manual labeling and there are fewer design decisions to make.
In a BWS setting, the amount of annotations needed to rank a given number of text instances depends on three variables, namely 1) corpus size (total number of texts used), 2) set size (number of texts in each comparison set), and 3) number of comparison sets each text appears in.

The number of comparisons divided by set size is regarded as the variable $N$, where $N=2$ generally yields good results in the literature \citep{kiritchenko-mohammad-2017-best}. A reliable set size is 4, since choosing the best and worst text instance from a 4-tuple set essentially provides the same number of comparisons
%results 
as five out of six possible pairwise comparisons (ibid).
%\citep{kiritchenko-mohammad-2017-best}.\todo{stimmt die citation hier? haben die das überhaupt gesagt?}

We take a random sample of 1,000 texts 
%from the data 
(excluding texts with \textsc{Absence} annotation, thus making the task harder, but giving us a more realistic distribution). With a set size of 4 and $N=2$, every text occurs in exactly 8 different sets and we get 2,000 comparison sets (tuples). These are then individually prompted to three LLMs: the relatively small Ministral-8B,\footnote{\url{https://huggingface.co/mistralai/Ministral-8B-Instruct-2410}} OpenAI's GPT-4 \citep{achiam2023gpt}, and the DeepSeek-V3 open source model \citep{liu2024deepseek}.

\begin{table}[htbp!]
\setlength{\aboverulesep}{1pt} % Reduce space above \midrule
\setlength{\belowrulesep}{2pt} % Reduce space below \midrule
\centering\scriptsize
\begin{tabular}{p{1.0cm}cccccccc}
\toprule
& \textbf{Annotator1} & \textbf{Annotator2} & \textbf{B}  & \textbf{W}  & \textbf{B + W} \\ \midrule
%\multicolumn{5}{c}{Models} \\
% Können wir die Tabelle bitte so lassen. Wir haben keinen Platz mehr. Habe es in der Caption erklärt.
\multirow{3}{*}{\parbox{1.0cm}{LLM-LLM}} & GPT4                  & DeepseekV3 & 0.73           & 0.69           & 0.56                  \\
& Ministral8B             & DeepseekV3 & 0.54           & 0.54           & 0.36                  \\ 
& GPT4                 & Ministral8B & 0.57           & 0.50           & 0.38                  \\ \midrule
\textbf{Average} &            &    & 0.61           & 0.57          & 0.43                \\ \midrule
%\multicolumn{5}{c}{Humans} \\
\multirow{5}{*}{\parbox{1.0cm}{Human-Human}} & AR                 & DS & 0.56           & 0.65           & 0.45                  \\ 
& DS                 & KB & 0.56           & 0.62           & 0.40                  \\   
& MR                 & AR & 0.51           & 0.65           & 0.39                 \\ 
& TP                 & AO & 0.73           & 0.55           & 0.48                 \\ 
& MP                 & MR & 0.59           & 0.52          & 0.41                 \\ 
\midrule
 \textbf{Average} &            &    & 0.59           & 0.60           & 0.43                 \\ \midrule 
%\multicolumn{5}{c}{Humans vs. Mistral8B} \\
\multirow{7}{*}{\parbox{1.0cm}{Human-LLM}} & AO                 & Ministral8B & 0.43           & 0.31           & 0.23                 \\ 
& AR                 & Ministral8B & 0.47           & 0.58           & 0.38                 \\ 
& DS                 & Ministral8B & 0.43           & 0.42           & 0.23                 \\
& KB                 & Ministral8B & 0.53           & 0.61           & 0.46                  \\ 
& MP                 & Ministral8B & 0.45           & 0.43           & 0.30                 \\
& MR                 & Ministral8B & 0.55           & 0.48           & 0.38                 \\ 
& TP                 & Ministral8B & 0.49           & 0.31           & 0.24                 \\
\midrule
\textbf{Average}       &     &    & 0.48           & 0.45          & 0.32                 \\ \midrule 
%\multicolumn{5}{c}{Humans vs. GPT4} \\
\multirow{7}{*}{\parbox{1.0cm}{Human-LLM}} & AO                & GPT4   & 0.68           & 0.55           & 0.45                 \\ 
& AR                  & GPT4 & 0.49           & 0.57           & 0.34                  \\ 
& DS                  & GPT4 & 0.44           & 0.71           & 0.43                \\ 
%GP                  & GPT4 & 0.44           & 0.49           & 0.28                  \\ 
& KB                  & GPT4 & 0.47           & 0.68           & 0.41                   \\ 
& MP                  & GPT4 & 0.57           & 0.62           & 0.41                   \\
& MR                  & GPT4 & 0.49           & 0.63           & 0.41                  \\ 
& TP                  & GPT4 & 0.63           & 0.57           & 0.43                  \\ 
\midrule 
 \textbf{Average} &             &      & 0.54           & 0.62           & 0.41 \\ \midrule
%AO                 & GP & 0.44           & 0.55           & 0.35                  \\  \\ \hline
%AO                 & GP & 0.44           & 0.55           & 0.35                  \\ 
%\multicolumn{5}{c}{Humans vs. DeepseekV3} \\
\multirow{7}{*}{\parbox{1.0cm}{Human-LLM}} & AO                 & DeepseekV3 & 0.61           & 0.59           & 0.45                 \\
& AR                 & DeepseekV3 & 0.55           & 0.68           & 0.41                \\
& DS                 & DeepseekV3 & 0.62           & 0.63           & 0.46                 \\ 
& KB                 & DeepseekV3 & 0.57           & 0.62           & 0.41                  \\   
& MP                 & DeepseekV3 & 0.69           & 0.53           & 0.41                 \\ 
& MR                 & DeepseekV3 & 0.59           & 0.68           & 0.46                 \\ 
& TP                 & DeepseekV3 & 0.58           & 0.58           & 0.41                \\ \midrule
 \textbf{Average} &            &    & 0.60           & 0.62           & 0.43                 \\ 
%\midrule
\bottomrule 
\end{tabular}%
\caption{Cohen's $\kappa$ Agreement between humans and LLMs in Best-Worst-Annotation (B: Best, W: Worst, B+W: Best + Worst). Two-letter shorthands for humans.}
\label{tab:bws_agreement}
\end{table}

Whereas Ministral-8B is run locally, we use the OpenAI API to access 
GPT-4 and the fireworks.ai API endpoint for DeepSeek-V3, since the DeepSeek-webservices are limited at the time of the experiment and hardware limitations hamper local deployment. Prompts are in the appendix.

We ask seven native German post-graduates to annotate one of two subsets of 50 tuples each with a custom browser-based annotation interface. Table~\ref{tab:bws_agreement} shows Cohen's $\kappa$ agreement across humans and LLMs.
We find that
agreement among humans is largely on par with agreement between humans and the two larger LLMs, while the lower agreement between Ministral-8B and humans, as well as the other machine annotators, indicates a limited capability of this model for the task at hand.
It appears that it is easier to identify the worst instance than the best, which is likely an artifact of our data. 
Interestingly, agreement between GPT-4 and DeepSeek-V3 is the highest overall, 
which could lend itself either to a) the task being easier for the LLMs than for humans, or b) that the models are overall fairly similar. We find no significant difference ($p=.118$) between GPT-4 and DeepSeek-V3 in Human-LLM comparison.

%human agreement is similar to the agreement between humans and GPT-4,\textcolor{blue}{ as well as between humans and DeepSeek-V3}, indicating i) that the task is overall feasable but by far not trivial, and ii) that GPT-4 \textcolor{blue}{and DeepSeek-V3}  are largely on par with humans. Furthermore, it appears that it is easier to identify the worst instance, rather than the best. \textcolor{blue}{Interestingly, agreement between GPT-4 and DeepSeek-V3 is the highest overall. On the other hand, the low agreement between Mistral-8B and human, as well as machine annotators indicates a limited capability of the model on the task at hand. } 

%the difficulty of the annotation, but also that the task is overal 

%, with agreements ranging from 0.44 - 0.71 Kappa,\todo{Durchschnittles Kappa angeben, zwischen Mensch vs. Maschine, und Mensch vs. Mensch!} which can be considered as moderate to good agreement \citep{Masson2003-gc}. 

\begin{equation}\scriptsize\label{eq:scaling}
    s(i) = \frac{\#best(i) - \#worst(i)}{\#overall(i)}
\end{equation}

By counting how often each text was chosen as the best, worst, or as one of two other texts, we calculate a score $s(i)$ as detailed in equation (\ref{eq:scaling}), resulting in an interval scale $[-1,1]$,
%incremented by $\frac{1}{8}$s, resulting in 17 equally spaced points,
which we normalize to a scale $[0,1]$. This scales (and ranks) the entire dataset, so it can be used for regression. 
It should be noted that the scores come in increments of $\frac{1}{8}$ (determined by number of comparisons of instance $i$), resulting in 17 discrete values.
%each instance is involved in), thus 
%at a scaling $[-1,1]$, 
%we get 17 discrete values.
We find a flat unimodal inverted U-shape in the score distribution without notable outliers. 
%is ranging from -1 to 1, scaling it to 0 to 1 after normalization. 

\subsubsection{Regression Models}

\iffalse
\todo{Fehlermeldung line 542, weiß nicht, wie man die weg bekommt. R2 A/B kursiv geht auch nicht weg}
\begin{table}
\centering\tiny
\begin{tabular}{|c|c|c|c|}
\toprule
 Features/Training Strategy & Model & MAE A/B    & R^2 A/B  \\ \hline\hline 
 Unigrams & KRR & 0.159/0.158 & 0.514/0.515 \\ 
Frozen LaBSE Embeddings & KRR & 0.118/0.117 & 0.678/0.686 \\
 Regression Head & bert-base-german & 0.149/0.158 & 0.516/0.490\\
Regression Head & LaBSE & 0.133/0.127 & 0.607/0.657 \\ %\hline
%Quantifier Lexicon + BSW         & Transfer Learning   & LaBSE & 0.162 & 0.474 \\ %\hline
%Mean values of Quantifiers + BSW & 
Reg. Head + Transfer   & LaBSE & 0.107/0.117 & 0.730/0.710 \\ \hline
\end{tabular}%
\caption{Comparison of different training strategies for 
%the optimization of 
regression based on BWS-Scaling. A: GPT-4 annotation, B: Deepseek-V3 annotation}
\label{tab:regression}
\end{table}
\fi

\begin{table*}
\centering
\footnotesize
\begin{tabular}{lccccc}
\toprule
 \textbf{Features/Training Strategy} & \textbf{Model} & \multicolumn{2}{c}{\textbf{MAE}} & \multicolumn{2}{c}{\textbf{R²}} \\  
 & & \textbf{GPT4} & \textbf{DSV3} & \textbf{GPT4} & \textbf{DSV3} \\ \midrule
 Unigrams & KRR & 0.159 & 0.158 & 0.514 & 0.515 \\ 
Frozen LaBSE Embeddings & KRR & 0.118 & 0.117 & 0.678 & 0.686 \\
 Regression Head & bert-base-german & 0.149 & 0.158 & 0.516 & 0.490\\
Regression Head & LaBSE & 0.133 & 0.127 & 0.607 & 0.657 \\ 
Reg. Head + Transfer & LaBSE & \textbf{0.107} & 0.117 & \textbf{0.730} & 0.710 \\ \bottomrule
\end{tabular}%
\caption{Comparison of different training strategies for regression based on BWS-Scaling. \\GPT4: GPT-4 BWS annotation, DSV3: Deepseek-V3 BWS annotation}
\label{tab:regression}
\end{table*}

We train 
%and evaluate 
a variety of different regression models with 5-fold cross validation
%, architectures and training strategies in order 
to optimize 
%regression 
%of the pairs of text labels 
for the values generated by Best-Worst Scaling, as shown in Table~\ref{tab:regression}.
%We applied 5-fold cross validation for the models in this section. 
We compare a Kernel Ridge Regression (KRR) baseline
%on unigrams and another on embedding features 
against 
%Sentence 
BERT-style-models
%\todo{bert-base ist kein sentence bert} 
with regression head, and test a transfer learning setup, for which we scale the 114 n-gram quantifiers as extracted from the binary Logistic Regression 
with another GPT-4 BWS,
%with GPT-4. 
then match these scores to the texts 
%in the training set 
and tune a LaBSE model on the same train/test split
before using it for the final task.
%It is curious that the 

Curiously, KRR with 
%frozen 
LaBSE embedding features benefits substantially from hyperparameter tuning, reaching superior results over LaBSE with regression head. The Transfer Model on GPT4 BWS offers the best performance, with acceptably high explained variance ($R^2=.73$) and only $.11$ Mean Absolute Error (MAE), which makes this model useful for downstream prediction as in the case study below. However, more data would likely also help, since training curves show continuous improvement.

\section{Case Study}\label{sec:discussion}

\begin{figure}[htbp!]
    \centering
    \includegraphics[width=.6\linewidth]{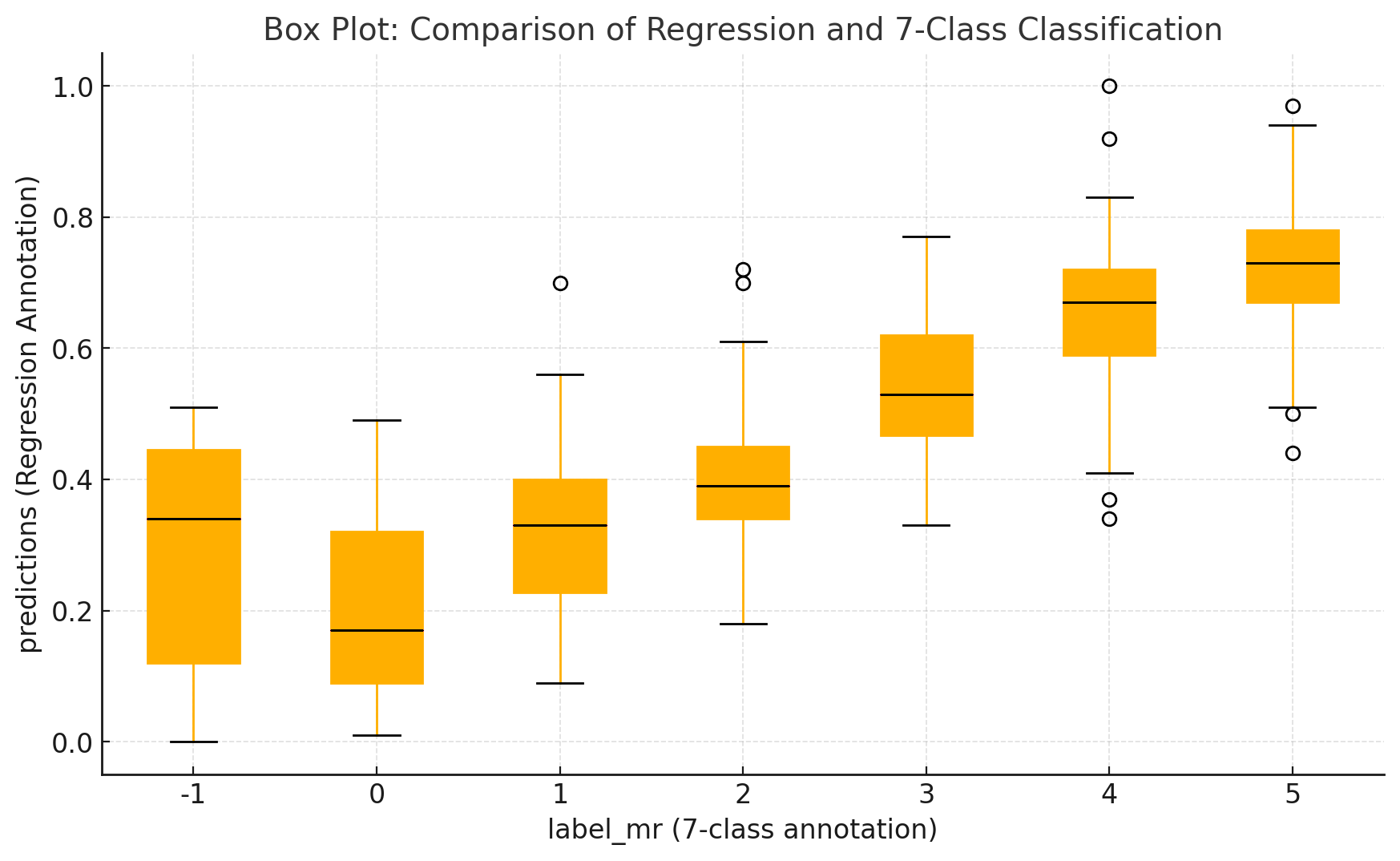}
    \caption{Multi-Class vs. Regression Distribution}
    \label{fig:multi-class-regression}
\end{figure}

For a proof of concept, we map the predictions of the regression model (LaBSE transfer regression model based on GPT-4 BWS) to the multi-class human annotation. Figure~\ref{fig:multi-class-regression} 
shows a strong relationship between multi-class labels and regression scores for the entire dataset (four species), but also that the extinction class is not properly represented in the regression, and furthermore that higher values are challenging to predict.

Figure~\ref{fig:rotwildotter} shows specie-specific distributions for Roe deer and Eurasian otter across all 119 offices, indicating a fairly good alignment between the regression result (top) and the multi-class annotation (bottom). 
However, the mapping is not unambiguous due to 1) shortcomings of the regression, such as the inability to model extinction and difficulty in predicting high values, and 2) imperfect alignment with class intervals, which are fuzzy with regard to the continuous values. However, pending further research, we find that our method performs well and produces plausible results.

%We conclude that the BWS-approach is robust and cost effective to estimate animal quantity. 

%This section details the case study, i.e., comparison of multi-class setup and regression.
%we used the data set annotated for multi-classification, containing all text entries of four selected species. 
%We let the best performing regression model mentioned in the previous section predict the dataset. A minor normalization of predicted values to [0;1] was conducted, after which we calculated the mean value of all the regressor's predictions for every class.\todo{das mean macht nicht so arg viel sinn. hier sollten die verteilungen respektive der Tierarten verglichen werden. Siehe Vergleich der in der Figure stattfindet. Was ist die Korrelation für Roe Deer, für Eurasian Otter, etc.} 

\begin{figure}[htbp!]
    \centering
    \includegraphics[width=.75\linewidth]{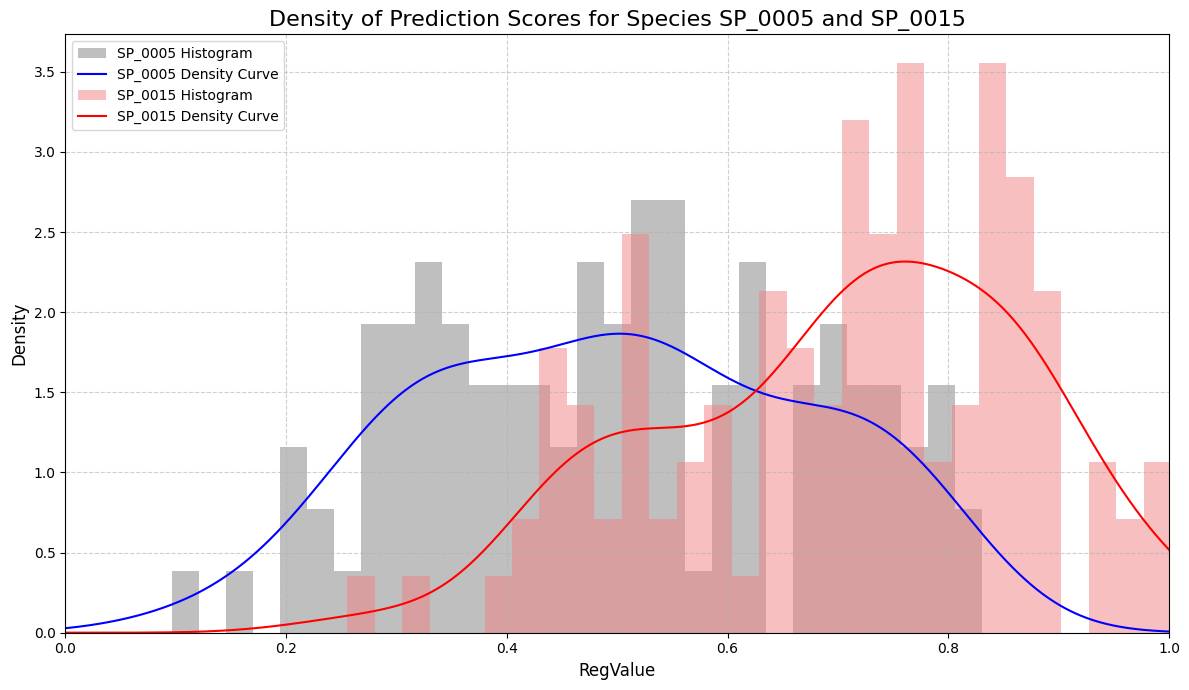}
    \includegraphics[width=.75\linewidth]{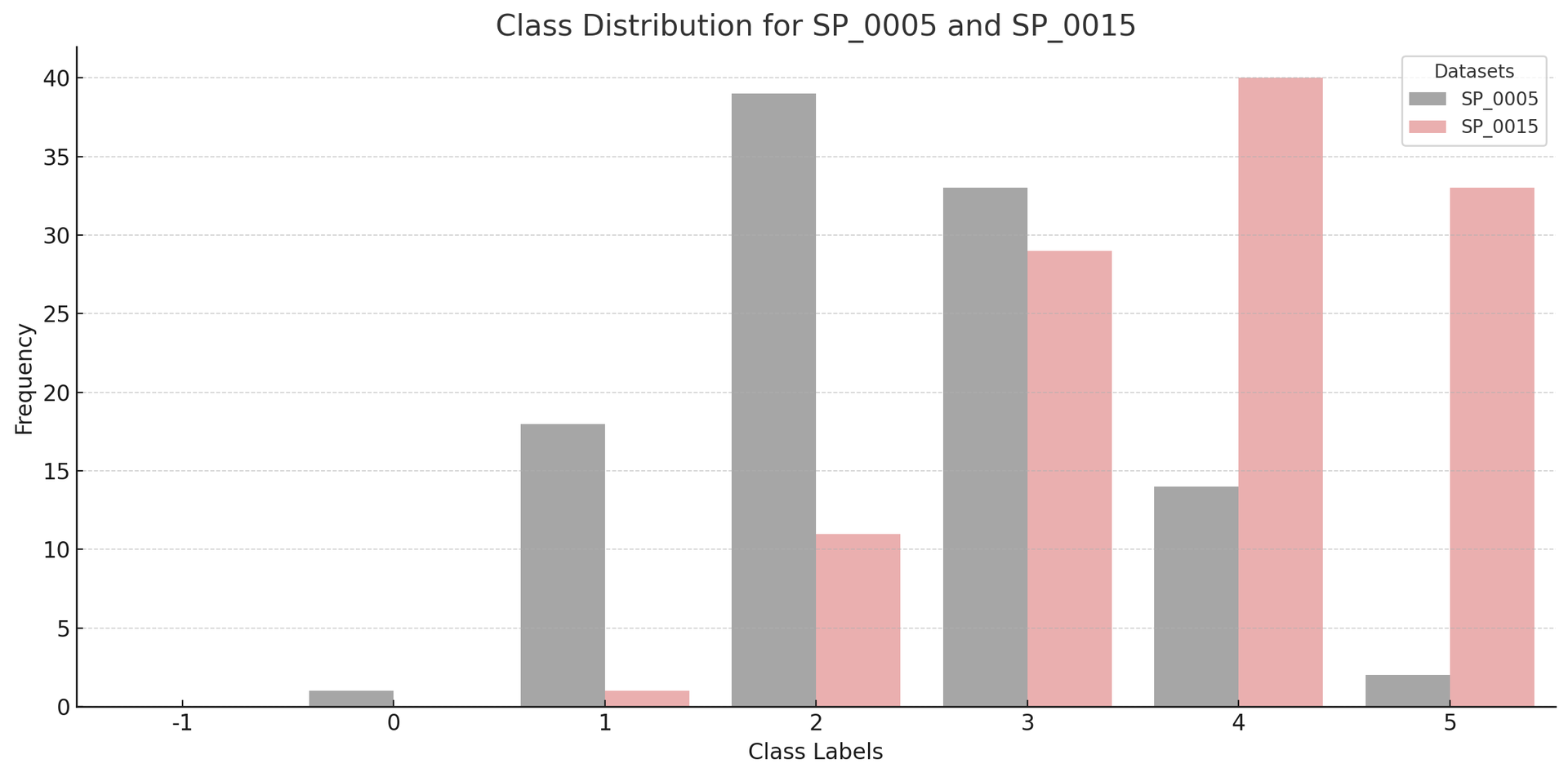}
    \caption{Density histogram of regressor prediction (top) and multi-class (bottom) distribution for Roe deer (SP\_0015, red) and Eurasian otter (SP\_0005, grey).}
    %across forestry offices as predicted by regression vs. multi-class annotation.}
    \label{fig:rotwildotter}
\end{figure}

\iffalse
\begin{figure}[htbp!]
    \centering
    \includegraphics[width=.65\linewidth]{figures/rotwild_fischotter_density.png}
    \includegraphics[width=.65\linewidth]{figures/multi_class_roedeer_otter.png}
    \caption{Density and histogram for Roe deer (SP\_0015, red) and Eurasian otter (SP\_0005, blue).}
    %across forestry offices as predicted by regression vs. multi-class annotation.}
    \label{fig:rotwildotter}
\end{figure}
\fi
%\clearpage

%\textcolor{red}{Tabelle \ref{tab:bws_agreement}: Das muss stark vereinfacht werden und im text erklärt. Außerdem Agreement zw. Menschen zeigen. Siehe zb Table 2 in \url{https://arxiv.org/pdf/2407.03916}}

%\hspace*{-0.5cm}
\section{Conclusion \& Future Work}\label{sec:conclusion}

This study demonstrates that information of occurrence frequencies from semi-structured historical biodiversity survey texts can be adequately modeled with Best-Worst Scaling through LLMs. %extracted using machine learning. 
While a simple classification approach performs well with minimal training data, a more complex classification struggles with design decisions and imbalanced data. BWS meets this by eliminating rating scale design decisions. In addition, it is cognitively and computationally less expensive, since no manual annotation of training data is necessary, while still offering similarly accurate results with much finer granularity through regression.
%Regression models estimate animal occurrences, reducing the need for manual annotation by leveraging large language models (LLMs). The Best-Worst-Scaling method created scalable training data, leading to effective regression models.
%Our dataset spans the spectrum between disaggregated and aggregated data (cf. Introduction). 

The robustness of methods and models should be further tested, not exclusive to biodiversity surveys, lending itself to a number of tasks. Yet, similar data to ours likely exists, e.g., on 19th century Bavarian flora, \citet{wikisource_oberamtsbeschreibungen}, or data in biodiversitylibrary.org, making our methods valuable.

\clearpage

\section*{Limitations}

%\textcolor{red}{Our best-worst-scaling approach is limited to data that can be scaled, thus it only applies to datasets that allow ranking, and that are diverse enough to result in a score distribution ...}

The accuracy of the method depends heavily on the capabilities of the specific LLM used. If a model lacks domain-specific knowledge or has biases, it may impact results.
Furthermore, without a reliable dataset to benchmark against, it is difficult to assess the absolute accuracy of the BWS-based regression approach, because we also test on BWS values. While we measured agreement on the BWS task with humans, it is impractical to scale the entire dataset with both LLMs and humans, and thus our agreement calculation may suffer from sampling bias. 

The effectiveness of the approach on different text sources or structured data remains uncertain. Differences in linguistic styles, terminologies, and data availability across domains may limit generalization. The approach assumes that frequency-related information in historical texts can be accurately mapped to numerical frequency estimates. If the original texts contain qualitative descriptions rather than explicit quantifiers, this may introduce errors. Also, older survey texts may reflect sampling biases, observer subjectivity, or incomplete data. If LLMs learn from these biases, the resulting quantity estimations may reinforce historical inaccuracies rather than correct them.

%Scalability and Computational Cost – While BWS reduces manual annotation needs, running large LLMs (like GPT-4) is still computationally expensive. The trade-offs between cost, accuracy, and granularity should be further evaluated.

\section*{Acknowledgements}

We thank our annotators from the Journal Club, and the colleagues in the Computational Historical Ecology group at the Chair for Computational Humanities at the University of Passau for their feedback.
%*******************
% BIBLIOGRAPHY
%*******************
\bibliographystyle{acl_natbib}
\bibliography{nodalida2025}

\begin{thebibliography}{26}
\expandafter\ifx\csname natexlab\endcsname\relax\def\natexlab#1{#1}\fi

\bibitem[{Achiam et~al.(2023)Achiam, Adler, Agarwal, Ahmad, Akkaya, Aleman, Almeida, Altenschmidt, Altman, Anadkat et~al.}]{achiam2023gpt}
Josh Achiam, Steven Adler, Sandhini Agarwal, Lama Ahmad, Ilge Akkaya, Florencia~Leoni Aleman, Diogo Almeida, Janko Altenschmidt, Sam Altman, Shyamal Anadkat, et~al. 2023.
\newblock Gpt-4 technical report.
\newblock \emph{arXiv preprint arXiv:2303.08774}.

\bibitem[{Bagdon et~al.(2024)Bagdon, Karmalkar, Gurulingappa, and Klinger}]{bagdon2024you}
Christopher Bagdon, Prathamesh Karmalkar, Harsha Gurulingappa, and Roman Klinger. 2024.
\newblock \href {https://doi.org/10.18653/v1/2024.naacl-long.439} {{\textquotedblleft}you are an expert annotator{\textquotedblright}: Automatic best{--}worst-scaling annotations for emotion intensity modeling}.
\newblock In \emph{Proceedings of the 2024 Conference of the North American Chapter of the Association for Computational Linguistics: Human Language Technologies (Volume 1: Long Papers)}, pages 7924--7936, Mexico City, Mexico. Association for Computational Linguistics.

\bibitem[{Berggren et~al.(2019)Berggren, Rama, and {\O}vrelid}]{johan-berggren-etal-2019-regression}
Stig~Johan Berggren, Taraka Rama, and Lilja {\O}vrelid. 2019.
\newblock \href {https://doi.org/10.18653/v1/W19-4409} {Regression or classification? automated essay scoring for {N}orwegian}.
\newblock In \emph{Proceedings of the Fourteenth Workshop on Innovative Use of NLP for Building Educational Applications}, pages 92--102, Florence, Italy. Association for Computational Linguistics.

\bibitem[{Br{\"o}nnimann et~al.(2018)Br{\"o}nnimann, Pfister, and White}]{Bronnimann2018-jg}
Stefan Br{\"o}nnimann, Christian Pfister, and Sam White. 2018.
\newblock Archives of nature and archives of societies.
\newblock In \emph{The Palgrave Handbook of Climate History}, pages 27--36. Palgrave Macmillan UK, London.

\bibitem[{Dorazio et~al.(2011)Dorazio, Gotelli, and Ellison}]{dorazio2011modern}
Robert~M Dorazio, Nicholas~J Gotelli, and Aaron~M Ellison. 2011.
\newblock Modern methods of estimating biodiversity from presence-absence surveys.
\newblock \emph{Biodiversity loss in a changing planet}, pages 277--302.

\bibitem[{Dornelas et~al.(2013)Dornelas, Magurran, Buckland, Chao, Chazdon, Colwell, Curtis, Gaston, Gotelli, Kosnik, McGill, McCune, Morlon, Mumby, Øvreås, Studeny, and Vellend}]{dornelas_biodiv}
Maria Dornelas, Anne~E. Magurran, Stephen~T. Buckland, Anne Chao, Robin~L. Chazdon, Robert~K. Colwell, Tom Curtis, Kevin~J. Gaston, Nicholas~J. Gotelli, Matthew~A. Kosnik, Brian McGill, Jenny~L. McCune, Hélène Morlon, Peter~J. Mumby, Lise Øvreås, Angelika Studeny, and Mark Vellend. 2013.
\newblock Quantifying temporal change in biodiversity: challenges and opportunities.
\newblock \emph{Proceedings of the Royal Society}, 280.

\bibitem[{Feng et~al.(2022)Feng, Yang, Cer, Arivazhagan, and Wang}]{feng2022languageagnosticbertsentenceembedding}
Fangxiaoyu Feng, Yinfei Yang, Daniel Cer, Naveen Arivazhagan, and Wei Wang. 2022.
\newblock \href {https://arxiv.org/abs/2007.01852} {Language-agnostic bert sentence embedding}.

\bibitem[{van Goethem and van Zanden(2021)}]{Van_Goethem2021-z}
Thomas van Goethem and Jan~Luiten van Zanden. 2021.
\newblock Biodiversity trends in a historical perspective.
\newblock In \emph{How Was Life? Volume II: New Perspectives on Well-being and Global Inequality since 1820}. Organisation for Economic Co-Operation and Development (OECD).

\bibitem[{Hoque and Sultana(2024)}]{hoque2024addressing}
Sk~Rezaul Hoque and Sk~Rima Sultana. 2024.
\newblock Addressing global environmental problems: Challenges, solutions, and opportunities.
\newblock \emph{The Social Science Review: A Multidisciplinary Journal}, 2(2):124--130.

\bibitem[{Iknayan et~al.(2014)Iknayan, Tingley, Furnas, and Beissinger}]{iknayan2014detecting}
Kelly~J Iknayan, Morgan~W Tingley, Brett~J Furnas, and Steven~R Beissinger. 2014.
\newblock Detecting diversity: emerging methods to estimate species diversity.
\newblock \emph{Trends in ecology \& evolution}, 29(2):97--106.

\bibitem[{Johnson et~al.(2011)Johnson, Brooks, Fenberg, Glover, James, Lister, Michel, Spencer, Todd, Valsami-Jones, Young, and Stewart}]{Johnson2011-ah}
Kenneth~G Johnson, Stephen~J Brooks, Phillip~B Fenberg, Adrian~G Glover, Karen~E James, Adrian~M Lister, Ellinor Michel, Mark Spencer, Jonathan~A Todd, Eugenia Valsami-Jones, Jeremy~R Young, and John~R Stewart. 2011.
\newblock Climate change and biosphere response: Unlocking the collections vault.
\newblock \emph{Bioscience}, 61(2):147--153.

\bibitem[{Kestemont et~al.(2022)Kestemont, Karsdorp, de~Bruijn, Driscoll, Kapitan, {\'O}~Mach{\'a}in, Sawyer, Sleiderink, and Chao}]{kestemont2022forgotten}
Mike Kestemont, Folgert Karsdorp, Elisabeth de~Bruijn, Matthew Driscoll, Katarzyna~A Kapitan, P{\'a}draig {\'O}~Mach{\'a}in, Daniel Sawyer, Remco Sleiderink, and Anne Chao. 2022.
\newblock Forgotten books: The application of unseen species models to the survival of culture.
\newblock \emph{Science}, 375(6582):765--769.

\bibitem[{Kiritchenko and Mohammad(2017)}]{kiritchenko-mohammad-2017-best}
Svetlana Kiritchenko and Saif Mohammad. 2017.
\newblock \href {https://doi.org/10.18653/v1/P17-2074} {Best-worst scaling more reliable than rating scales: A case study on sentiment intensity annotation}.
\newblock In \emph{Proceedings of the 55th Annual Meeting of the Association for Computational Linguistics (Volume 2: Short Papers)}, pages 465--470, Vancouver, Canada. Association for Computational Linguistics.

\bibitem[{Kommineni et~al.(2024)Kommineni, Ahmed, Koenig-Ries, and Samuel}]{Kommineni2024-ne}
Vamsi~Krishna Kommineni, Waqas Ahmed, Birgitta Koenig-Ries, and Sheeba Samuel. 2024.
\newblock Automating information retrieval from biodiversity literature using large language models: A case study.
\newblock \emph{Biodivers. Inf. Sci. Stand.}, 8.

\bibitem[{K{\"o}nig et~al.(2019)K{\"o}nig, Weigelt, Schrader, Taylor, Kattge, and Kreft}]{konig2019biodiversity}
Christian K{\"o}nig, Patrick Weigelt, Julian Schrader, Amanda Taylor, Jens Kattge, and Holger Kreft. 2019.
\newblock Biodiversity data integration—the significance of data resolution and domain.
\newblock \emph{PLoS biology}, 17(3):e3000183.

\bibitem[{Langer et~al.(2024)Langer, Burghardt, Borgards, Richter, and Wirth}]{langer2024relation}
Lars Langer, Manuel Burghardt, Roland Borgards, Ronny Richter, and Christian Wirth. 2024.
\newblock The relation between biodiversity in literature and social and spatial situation of authors: Reflections on the nature--culture entanglement.
\newblock \emph{People and Nature}, 6(1):54--74.

\bibitem[{Likert(1932)}]{likert1932technique}
Rensis Likert. 1932.
\newblock A technique for the measurement of attitudes.
\newblock \emph{Archives of Psychology}.

\bibitem[{Liu et~al.(2024)Liu, Feng, Xue, Wang, Wu, Lu, Zhao, Deng, Zhang, Ruan et~al.}]{liu2024deepseek}
Aixin Liu, Bei Feng, Bing Xue, Bingxuan Wang, Bochao Wu, Chengda Lu, Chenggang Zhao, Chengqi Deng, Chenyu Zhang, Chong Ruan, et~al. 2024.
\newblock Deepseek-v3 technical report.
\newblock \emph{arXiv preprint arXiv:2412.19437}.

\bibitem[{L{\"u}cking et~al.(2022)L{\"u}cking, Driller, Stoeckel, Abrami, Pachzelt, and Mehler}]{lucking2022multiple}
Andy L{\"u}cking, Christine Driller, Manuel Stoeckel, Giuseppe Abrami, Adrian Pachzelt, and Alexander Mehler. 2022.
\newblock Multiple annotation for biodiversity: developing an annotation framework among biology, linguistics and text technology.
\newblock \emph{Language resources and evaluation}, 56(3):807--855.

\bibitem[{Qian et~al.(2021)Qian, Feng, Wen, Ma, and Xie}]{qian-etal-2021-counterfactual}
Chen Qian, Fuli Feng, Lijie Wen, Chunping Ma, and Pengjun Xie. 2021.
\newblock \href {https://doi.org/10.18653/v1/2021.acl-long.422} {Counterfactual inference for text classification debiasing}.
\newblock In \emph{Proceedings of the 59th Annual Meeting of the Association for Computational Linguistics and the 11th International Joint Conference on Natural Language Processing (Volume 1: Long Papers)}, pages 5434--5445, Online. Association for Computational Linguistics.

\bibitem[{Rehbein et~al.(2024)Rehbein, Escobari~Vargas, Fischer, Güntsch, Haas, Matheisen, Perschl, Wieshuber, and Engel}]{rehbein2024historical}
Malte Rehbein, Andrea~Belen Escobari~Vargas, Sarah Fischer, Anton Güntsch, Bettina Haas, Giada Matheisen, Tobias Perschl, Alois Wieshuber, and Thore Engel. 2024.
\newblock \href {https://doi.org/10.5281/zenodo.14008158} {Historical animal observation records by bavarian forestry offices (1845): Description of the data sets}.
\newblock Version 1.3 as of 2024-10-29, \url{https://doi.org/10.5281/zenodo.14008158}.

\bibitem[{Ribeiro et~al.(2016)Ribeiro, Singh, and Guestrin}]{ribeiro2016should}
Marco~Tulio Ribeiro, Sameer Singh, and Carlos Guestrin. 2016.
\newblock " why should i trust you?" explaining the predictions of any classifier.
\newblock In \emph{Proceedings of the 22nd ACM SIGKDD international conference on knowledge discovery and data mining}, pages 1135--1144.

\bibitem[{Schuman and Presser(1996)}]{schuman1996questions}
Howard Schuman and Stanley Presser. 1996.
\newblock \emph{Questions and answers in attitude surveys: Experiments on question form, wording, and context}.
\newblock Sage.

\bibitem[{Telenius(2011)}]{telenius2011biodiversity}
Anders Telenius. 2011.
\newblock Biodiversity information goes public: Gbif at your service.
\newblock \emph{Nordic Journal of Botany}, 29(3):378--381.

\bibitem[{Wang and Culotta(2020)}]{wang-culotta-2020-identifying}
Zhao Wang and Aron Culotta. 2020.
\newblock \href {https://doi.org/10.18653/v1/2020.findings-emnlp.308} {Identifying spurious correlations for robust text classification}.
\newblock In \emph{Findings of the Association for Computational Linguistics: EMNLP 2020}, pages 3431--3440, Online. Association for Computational Linguistics.

\bibitem[{{Württembergische Oberamtsbeschreibungen}(1824--1886)}]{wikisource_oberamtsbeschreibungen}
{Württembergische Oberamtsbeschreibungen}. 1824--1886.
\newblock \href {https://de.wikisource.org/\\wiki/Württembergische\_Oberamtsbeschreibungen} {{Wikisource}}.
\newblock Accessed: 30 Jan. 2025.

\end{thebibliography}

%\clearpage 

\section*{APPENDIX: PROMPTS}\label{sec:appendix}

%\subsection*{GPT-4 Prompts}
\begin{small}
\subsection*{Multi-Classification Prompt}

    \paragraph{System-prompt:}
    You are a German native expert in text classification. Use the provided classification scheme to classify German texts based on species frequency descriptions.
    
    \paragraph{User-prompt:}
    You are a classification model. Classify the given German text into one of the following categories:\\
- Abundant (5): Species is very frequently observed or present.\\
- Common (4): Species is commonly found in the area.\\
- Common to Rare (3): Species is observed, but not very frequently.\\
- Rare (2): Species is rarely seen in the area.\\
- Very Rare (1): Species is seen only in exceptional circumstances.\\
- Absent (0): Species is not observed in the area.\\
- Extinct (-1): Species no longer exists in the area.\\ Read the provided text and classify it according to this scheme.
    Here is the text to classify:\\
    Text

\subsection*{Best-Worst Scaling Prompt}

    \paragraph{System-prompt:} 
    You are an expert annotator specializing in Best-Worst Scaling 
    of German texts based on quantity information about animal occurrences.

    \paragraph{User-prompt:}
(Texts 1 to 4 were substituted with the actual texts of a tuple): Task: From the following German texts about animal occurrence, identify:\\ Best: The text conveying the highest quantity (e.g., presence, frequency, population size)\\ Worst: The text conveying the lowest quantity. \\1. Text 1 \\2. Text 2 \\3. Text 3\\4. Text 4\\ JSON format for your answer:\\\{ "Best": [Text Number],\\  "Worst": [Text Number]\}
\end{small}

\end{document}